\documentclass[cameraready]{Interspeech}
\title{Uncovering Latent Depression Severity for Binary Depression Detection via Advantage-weighting Ranking}

\author[affiliation={1}, orcid=0009-0008-4810-7428]{Manning}{Gao}
\author[affiliation={1}]{Tingyi}{Liu}
\author[affiliation={1}, orcid=0009-0006-1648-4873]{Leheng}{Zhang}
\author[affiliation={2}, orcid=0000-0002-4884-323X]{Haifeng}{Hu}
\author[affiliation={1}]{Yuncheng}{Jiang}
\author[affiliation={1}, orcid=0000-0001-9763-375X, correspondingauthor]{Sijie}{Mai}
\address{
    $^1$ South China Normal University, China \
    $^2$ Sun Yat-sen University, China \
}
\email{20232005149@m.scnu.edu.cn, liutingyi@m.scnu.edu.cn, lehengzhang@m.scnu.edu.cn, huhaif@mail.sysu.edu.cn, ycjiang@scnu.edu.cn, sijiemai@m.scnu.edu.cn}


\usepackage{comment}
\usepackage{multirow}
\usepackage{subcaption}
\usepackage{graphicx}
\usepackage{booktabs}  
\usepackage{amsmath}
\usepackage{cite}
\usepackage{siunitx}
\begin{document}

\maketitle

\begin{abstract}
    Automatic depression detection using audio-visual data faces significant challenges, particularly in disentangling overlapping feature distributions and establishing robust decision boundaries. To address this, we propose a fine-grained multimodal framework featuring a temporal encoder and a mutual transformer to facilitate deep cross-modal fusion. Our core contribution is the Binary Advantage-weighting Ranking Loss, which optimizes the latent space distribution through two complementary mechanisms: Advantage-weighted Separation, which mines hard pairs by computing a pairwise prediction difference matrix and dynamically weighting them based on their difficulty; and Advantage-weighted Compactness, which minimizes intra-class variance to force features to cluster around their respective class centers. Extensive experiments on D-vlog and LMVD demonstrate that our model reconstructs the latent ordinal structure by prioritizing hard pairs, thereby achieving state-of-the-art performance.
\end{abstract}

\section{Introduction and Related Work}
Depression is a global mental health challenge. Automatic Depression Detection (ADD) using multimodal data (e.g., audio, visual)~\cite{10843708} has emerged as a promising non-invasive screening tool. While deep learning has improved feature extraction, detecting depression in user-generated content (e.g., vlogs) remains challenging due to the subtle and ambiguous boundaries between depressed and non-depressed behaviors. To capture these intricate temporal patterns, Transformer-based models~\cite{liu2025depformer, HE2026103632} have been widely adopted for their capability to model global dependencies through self-attention mechanisms. Furthermore, to fully leverage the complementary information across different streams, recent studies have emphasized deep cross-modal interactions, such as employing cross-attention scaling layers coupled with advanced tensor-based pooling methods to fuse multimodal representations~\cite{ilias24_interspeech}. More recently, Mamba-based architectures~\cite{10889975, lin2025ste, 11465117} have emerged as efficient alternatives, leveraging selective state space models to handle long sequences with linear computational complexity.

Beyond architectural innovation, a critical bottleneck lies in optimization objectives. Most current approaches~\cite{11223210, 10889975, LI2026115423, SHABANA2026108561} rely on pointwise supervision (e.g., Binary Cross-Entropy, BCE), which treats depressed and non-depressed samples as independent nominal classes. This formulation neglects the latent ordinal nature of depression severity, a continuous spectrum in which severe cases should be ranked higher than mild or normal ones~\cite{Zuo2025LeveragingOI, 8913383, jayawardena2020ordinal, 6883166}. Although vlog datasets provide only discrete binary annotations, the underlying physiological and behavioral symptoms exist on a continuous scale. To bridge this gap, we adopt a pairwise learning paradigm~\cite{10.1145/3696410.3714841}. By encouraging the model to learn a fine-grained ranking of depression risk, this approach effectively reconstructs the intrinsic ordinal relationships from coarse binary labels. Integrating ordinal learning into ADD is complicated by standard datasets that offer only binary labels rather than fine-grained severity scores. The resulting sparse supervision~\cite{Zuo2025LeveragingOI} hinders precise ranking inference. Furthermore, standard pairwise losses in affective computing typically treat all pairs equally. The uniform weighting is insufficient for vlog datasets where feature distributions are highly overlapping (e.g., depressed individuals hiding symptoms often share similar audio-visual cues with normal controls). A standard loss that averages easy and hard pairs fails to enforce a clear decision boundary in these ambiguous regions.

To address these aforementioned problems, we propose a Binary Advantage-weighting Ranking (BAR) Loss that builds upon Mutual Transformer-based multimodal fusion~\cite{11342588}. Unlike recent methods that perform hard instance mining by masking salient features within a single sample~\cite{zhou25f_interspeech}, our method addresses inter-sample ambiguity by formulating depression detection as a ranking problem to maximize the separation between positive and negative groups. The core innovation is an error-aware advantage-weighting mechanism that dynamically assigns higher weights to ambiguous hard pairs that violate the decision boundary. Driven by this mechanism, the BAR Loss imposes dual geometric constraints: it enforces inter-class separation (via a hinge margin) to establish a strict ranking order, while simultaneously promoting intra-class compactness (via variance reduction) to mitigate the issue of feature overlap. Our main contributions are as follows:
\begin{itemize}
    % \item We propose a multimodal architecture that achieves more fine-grained and effective multimodal fusion than baselines.
    \item We introduce a pairwise learning paradigm to ADD, recovering the latent continuous spectrum of depression that is typically obscured by coarse binary labels.
    \item We design the BAR Loss, which incorporates a dynamic advantage-weighting mechanism to prioritize hard pairs, effectively solving the challenges of sparse binary supervision and high feature overlap.
\end{itemize}

\section{Methods}
\begin{figure*}[t]
  \centering
  \vspace{-0.6em}
  \includegraphics[width=0.85\linewidth]{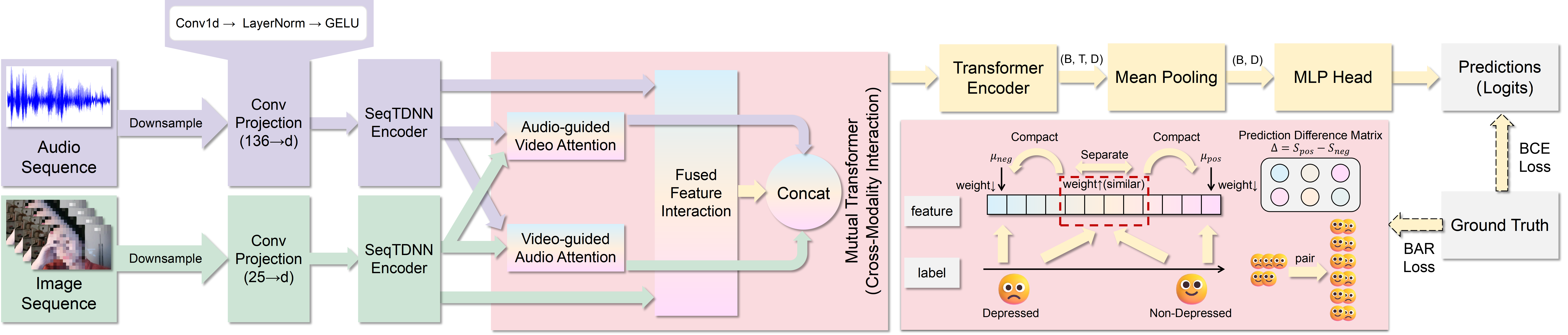}
  \caption{The forward pass model and core loss calculation.}
  \vspace{-0.6em}
  \label{fig:model}
\end{figure*}
Let the dataset be denoted as $\mathcal{D} = \{ (X_a^i, X_v^i, y^i) \}_{i=1}^N$, where $X_a \in \mathbb{R}^{T \times d_a}$ and $X_v \in \mathbb{R}^{T \times d_v}$ represent the acoustic and visual feature sequences with length $T$, and $y^i \in \{0, 1\}$ is the binary label for depression detection.
\subsection{Unimodal Temporal Encoding} 
To capture local temporal dependencies and align feature dimensions, we employ a dual-stream encoder consisting of 1D Convolutional Projections and Sequence Time-Delay Neural Networks (Seq-TDNN)~\cite{Desplanques2020}. For a modality $m \in \{a, v\}$, denoting audio and video modalities, the input sequence $X_m$ is first subjected to a projection to a hidden dimension $d_h$:
\vspace{-0.4em}
\begin{align}
H_m^{(0)} = \text{GELU}(\text{LayerNorm}(\text{Conv1D}(X_m)))   
\end{align}
Subsequently, the features pass through the Seq-TDNN encoder to capture long-range temporal contexts, yielding the encoded representations $H_a, H_v \in \mathbb{R}^{T' \times d_h}$.

\subsection{Mutual Transformer Fusion}
To model the interplay between acoustic and visual modalities, we utilize a Mutual Transformer followed by a fusion encoder.

\noindent \textbf{Cross-Modal Interaction:} Before cross-modal fusion, the unimodal representations $H_a$ and $H_v$ are further refined through modality-specific Transformer encoders to obtain $E_a$ and $E_v$. We then compute three distinct attention streams to capture bidirectional and joint interactions. For instance, the mutual attention from audio to video (where audio acts as the query) is computed by deriving $Q_a$ from $E_a$, and $K_v, V_v$ from $E_v$:
\vspace{-0.4em}
\begin{align}
    \text{Attention}_{a \rightarrow v} = \text{Softmax}\left(\frac{Q_a K_v^T}{\sqrt{d_k}}\right) V_v
\end{align}
Symmetrically, we compute $\text{Attention}_{v \rightarrow a}$. Additionally, a joint self-attention stream $\text{Attention}_{f \rightarrow f}$ is calculated based on the concatenated features $F = [E_a, E_v]$. These streams are concatenated along the feature dimension to form a comprehensive cross-modal representation, which is then passed through an $l$-layer Transformer Encoder to obtain the final fused sequence.

\noindent \textbf{Prediction:} The final representation $z$ is obtained by mean-pooling the final fused sequence. A Multi-Layer Perceptron (MLP) predicts the depression score $s \in \mathbb{R}$, incorporating an initial dropout layer for regularization:
\vspace{-0.4em}
\begin{align}
    h &= \text{ReLU}(W_1 \text{Dropout}(z) + b_1) \\ s &= W_2 \text{Dropout}(h) + b_2
\end{align}
The probability of depression is given by $p = \sigma(s)$, where $\sigma$ is the sigmoid function.

\noindent \textbf{Dynamic Thresholding: }The BAR Loss explicitly imposes constraints of margin separation and intra-class compactness on the predictions. This optimization fundamentally reshapes the score distribution, potentially causing the optimal separating hyperplane to deviate from the default median. 

To calibrate the decision boundary, we adopt a dynamic thresholding strategy~\cite{10.1007/978-3-662-44851-9_15, 10.5555/1597538.1597615} during the inference phase. Let $\mathcal{D}_{val} = \{(X_i, y_i)\}$ denote the validation set. We perform a grid search for the optimal threshold $\tau^*$ that maximizes a specific evaluation metric $\mathcal{M}$ (e.g., F1):
\vspace{-0.4em}
\begin{align}
    \tau^* = \operatorname*{argmax}_{\tau \in [0, 1]} \mathcal{M}\left(y_{val}, \mathbb{I}(p_{val} > \tau)\right)
\end{align}
where $p_{val}$ represents the predicted probabilities on the validation set, and $\mathbb{I}(\cdot)$ is the indicator function. The final prediction for a test sample is determined by $\hat{y}_{test} = \mathbb{I}(p_{test} > \tau^*)$, using the learned distribution instead of an arbitrary boundary.

\subsection{Binary Advantage-weighting Ranking Loss} 
To address the distribution overlap between depressed and non-depressed samples, we propose the BAR Loss, which optimizes the decision boundary by mining hard pairs and regularizing the distribution of prediction scores.

\noindent {\textbf{Separation:} }Let $\mathcal{S}_{pos} = \{s_i | y_i=1\}$ and $\mathcal{S}_{neg} = \{s_j | y_j=0\}$ be the sets of prediction scores for positive and negative samples in a batch. We construct a pairwise difference matrix $D_{ij} = s_i - s_j$. The relative difficulty matrix $R$ is defined as:
\vspace{-0.4em}
\begin{align}
    R_{ij} = \sigma(D_{ij}), \quad \forall s_i \in \mathcal{S}_{pos}, s_j \in \mathcal{S}_{neg}
\end{align}
We normalize $R$ to obtain the advantage matrix $A$:
\vspace{-0.4em}
\begin{align}
    A_{ij} = \frac{R_{ij} - \mu_R}{\text{std}(R) + \epsilon}
\end{align}
where $\mu_R$ and $\text{std}(R)$ are the mean and standard deviation of $R$, and $\epsilon$ (e.g., $10^{-6}$) prevents division by zero. For numerical stability, we clip $A_{ij}$ to the range $[-10, 10]$ before computing the weights. The weighting matrix $W$ highlights hard pairs (where the model is less confident or wrong) and is computed as:
\vspace{-0.4em}
\begin{align}
    W_{ij} = \text{ReLU}(-A_{ij})
\end{align}
The separation loss forces a margin $m$ between positive and negative scores, weighted by their difficulty:
\vspace{-0.4em}
\begin{align}
    \mathcal{L}_{sep} = \frac{\sum_{i,j} W_{ij} \cdot \max(0, m - (s_i - s_j))}{\sum_{i,j} W_{ij} + \epsilon}
\end{align}

\noindent \textbf{Compactness:} We minimize score distances to their own class means ($\mu_{pos}, \mu_{neg}$), weighted by the aggregated advantage weights $\bar{W}$, reducing intra-class variance:
\vspace{-0.4em}
\begin{align}
    \mathcal{L}_{com} = \frac{1}{2} \sum_{k \in \{pos, neg\}} \frac{1}{|\mathcal{S}_k|} \sum_{s \in \mathcal{S}_k} (1 + \beta \bar{W}_s) (s - \mu_k)^2
\end{align}
where $\beta$ controls the strength of mining hard pairs.

\noindent \textbf{Distribution Regularization:} To prevent representation collapse, we constrain the batch mean ($\mu_p$) and standard deviation ($\sigma_p$) of predicted probabilities to prior targets $\mu_{target}=0.5$ and $\sigma_{target}=0.5$, which serve as neutral anchors for a balanced binary distribution and encourage ideal entropy by pushing sigmoid predictions toward 0 and 1.
\vspace{-0.4em}
\begin{align}
    \Delta\mu = \mu_p - \mu_{target},\quad \Delta\sigma = \sigma_p - \sigma_{target}
\end{align}
Distribution Regularization Loss is calculated as:
\vspace{-0.4em}
\begin{align}
\mathcal{L}_{reg} = \max(0, \Delta\mu^2 - \gamma) + \max(0, \Delta\sigma^2 - \gamma)
\end{align}
where $\gamma$ is a relaxation margin.

\noindent \textbf{Total Loss:} It is a weighted sum of $\mathcal{L}_{BCE}$ and $\mathcal{L}_{BAR}$:
\begin{align}
    \mathcal{L}_{t} = \mathcal{L}_{BCE} + \lambda_{BAR}(\mathcal{L}_{sep} + \lambda_{com} \mathcal{L}_{com} + \lambda_{reg} \mathcal{L}_{reg})
\end{align}

\section{Experiments, Results, and Discussion}
\subsection{Datasets}
To evaluate the performance of our proposed method in real-world scenarios, we utilized two large-scale multimodal vlog datasets. Unlike traditional laboratory-controlled datasets, both consist of videos collected in the wild, capturing spontaneous non-verbal behaviors of individuals in their daily lives.
\begin{itemize}
    \item D-vlog~\cite{Yoon_Kang_Kim_Han_2022}: This dataset comprises 961 vlogs (approximately 160 hours) collected from YouTube, featuring 816 distinct speakers. The samples are categorized based on self-reported depression symptoms within the vlogs, resulting in 555 depressed and 406 non-depressed videos.
    \item LMVD~\cite{HE2026103632}: This dataset contains 1,823 video samples (approximately 214 hours) collected from four major multimedia platforms: Sina Weibo, Bilibili, TikTok, and YouTube. It includes data from 1,475 participants, balanced between 908 depressed and 915 non-depressed samples, annotated by volunteers and verified by clinicians.
\end{itemize}
\subsection{Implementation Details}
\begin{figure}[t]
    \centering
    \includegraphics[width=0.65\linewidth]{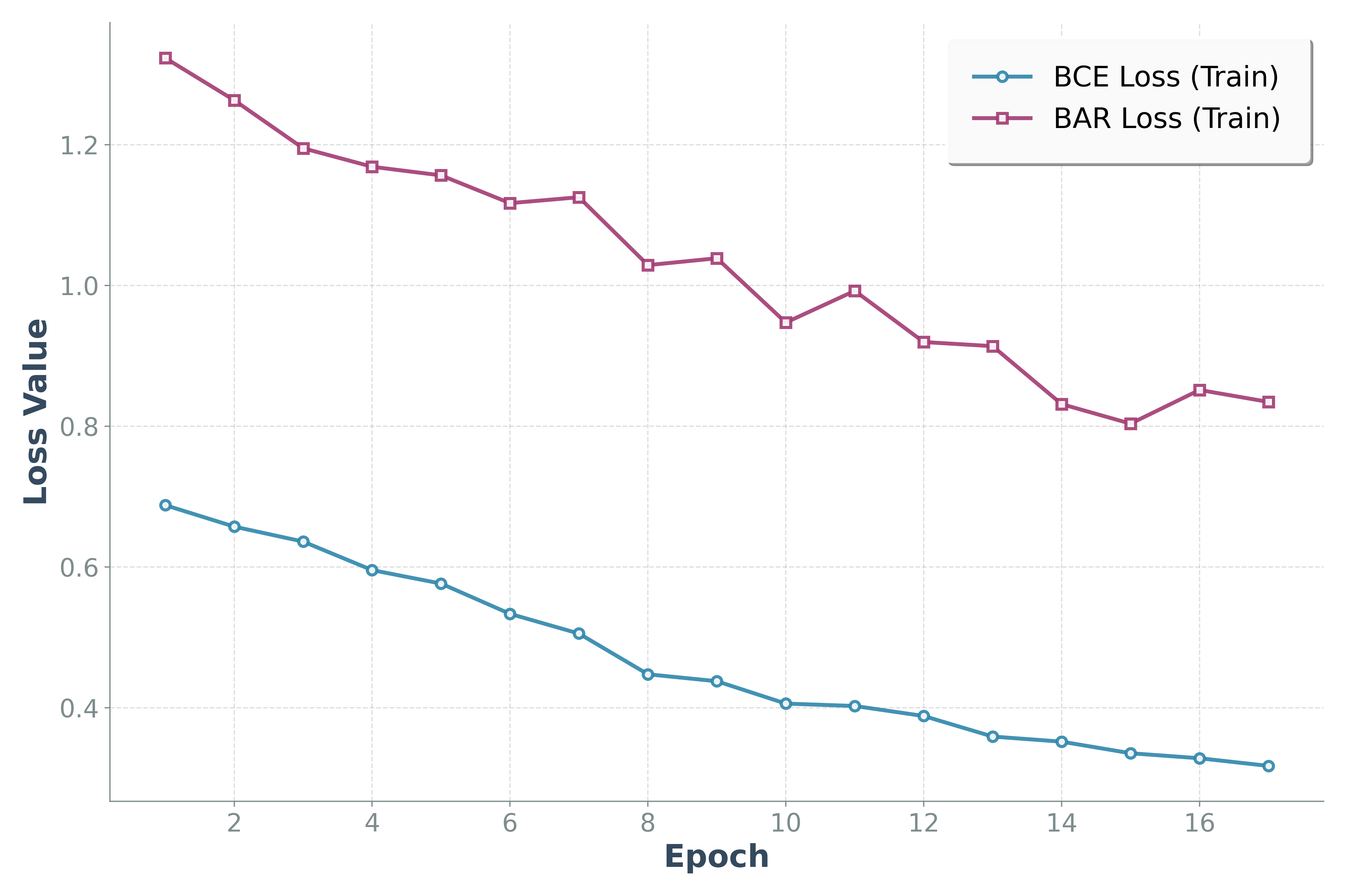}
    \caption{Training loss curves on LMVD.}
    \vspace{-0.5em}
    \label{fig:loss}
\end{figure}

Based on Optuna~\cite{DBLP:journals/corr/abs-1907-10902}, we defined a reasonable range and applied a random search with a limited number of trials (50) to optimize the hyperparameters and set the optimization objective as F1. Detailed hyperparameters are shown in Table~\ref{tab:hyperparams_comparison}. Training loss curves of BCE and BAR Loss are shown in Figure~\ref{fig:loss}.
\begin{table}[t]
\centering
\small
\caption{Comparison of hyperparameters on different datasets. Values are rounded to three significant figures for clarity.}
\label{tab:hyperparams_comparison}
\setlength{\tabcolsep}{4pt}
\begin{tabular}{@{}llll@{}}
\noalign{\hrule height 1pt}
\textbf{Parameter} & \textbf{D-vlog} & \textbf{LMVD}\\
\midrule
Optimizer & AdamW & AdamW \\
Batch Size & \num{32} & \num{16} \\
Seed & \num{123} & \num{123} \\
Epochs & \num{150} & \num{100} \\
Learning Rate & \num{4.65e-5} & \num{1.76e-5}\\
Dropout & \num{3.07e-1} & \num{4.18e-1}\\
Weight Decay & \num{2.51e-4} & \num{4.81e-3} \\
Fully Connected Layer Size & \num{256} & \num{512} \\
Separation Loss Margin ($m$) & \num{4.06e-1} & \num{1.15e0} \\
Compactness Loss Weight ($\lambda_{com}$) & \num{4.89e-2} & \num{4.51e-2} \\
Advantage-weighting Beta ($\beta$) & \num{7.60e-1} & \num{1.03e0} \\
BAR Loss Weight ($\lambda_{BAR}$) & \num{8.39e-1} & \num{5.61e-1} \\
Regularization Loss Weight ($\lambda_{reg}$) & \num{7.16e-2} & \num{2.21e-4} \\
Regularization Loss Margin ($\gamma$) & \num{4.04e-2} & \num{3.76e-2} \\
\noalign{\hrule height 1pt}
\end{tabular}
\end{table}

\subsection{Main Results}
Table~\ref{tab:performance_comparison} presents the quantitative comparison with baselines. On the LMVD dataset, our proposed framework outperforms all baselines, achieving the highest results across all metrics (F1: 77.01). On the D-vlog dataset, it also demonstrates superior performance, securing the best metrics (F1: 77.66). Although DepMamba~\cite{10889975} achieves a marginally higher recall, we maintain a better balance between precision and recall, resulting in a superior overall F1. Results validate the effectiveness and robustness of our framework across diverse data distributions.
\begin{table*}[ht]
\centering
\small
\caption{Performance Comparison of Different Methods. The best results are in bold and the second best results are underlined.}
\label{tab:performance_comparison}
\begin{tabular}{lcccccccccc}
\noalign{\hrule height 1pt}
\multirow{2}{*}{\textbf{Methods}} & \multicolumn{5}{c}{\textbf{D-vlog}} & \multicolumn{5}{c}{\textbf{LMVD}} \\
\cmidrule(lr){2-6} 
\cmidrule(lr){7-11}
& \textbf{Accuracy} & \textbf{Precision} & \textbf{Recall} & \textbf{F1} & \textbf{Average} & \textbf{Accuracy} & \textbf{Precision} & \textbf{Recall} & \textbf{F1} & \textbf{Average} \\ 
\midrule
% KNN~\cite{pampouchidou2017facial}       & 60.38 & 61.41 & 85.64 & 71.47 & 69.73 & 56.83 & 57.43 & 50.92 & 53.92 & 54.78 \\
Bi-LSTM~\cite{10.1145/3347320.3357696}   & 64.47 & 67.68 & 75.34 & 71.10 & 69.65 & 66.85 & 65.81 & 70.33 & 67.83 & 67.71 \\
TBN~\cite{9010900}       & 63.21 & 69.99 & 65.31 & 67.16 & 66.42 & 67.94 & 67.06 & 70.33 & 68.56 & 68.47 \\
STST~\cite{TAO2024577}      & 61.79 & 64.67 & 77.51 & 69.77 & 68.44 & 67.76 & 69.20 & 64.01 & 66.23 & 66.80 \\
% TFN~\cite{zadeh-etal-2017-tensor}       & 67.14 & 72.38 & 70.19 & 71.23 & 70.24 & 63.93 & 64.08 & 62.64 & 63.34 & 63.50 \\
DepTrans~\cite{Yoon_Kang_Kim_Han_2022}  & 62.89 & 64.43 & 84.82 & 72.54 & 71.17 & 61.93 & 60.36 & 72.16 & 65.08 & 64.88 \\
TAMFN~\cite{9961146}     & 67.45 & 68.08 & 82.93 & 74.75 & 73.30 & 70.49 & 71.15 & 68.86 & 69.84 & 70.09 \\
STE-Mamba~\cite{lin2025ste}     & 69.34 & 68.35 & \underline{86.80} & 76.49 & 75.25 & 71.58 & 68.57 & \textbf{79.12} & 73.47 & 73.19 \\
%DepMamba~\cite{10889975}       & \underline{68.87} & \underline{68.19} & \textbf{86.99} & \underline{76.44} & \underline{75.12} & \underline{72.13} & \underline{70.18} & \underline{76.56} & \underline{73.20} & \underline{73.02} \\ 
DepMamba~\cite{10889975}       & 68.87 & 68.19 & \textbf{86.99} & 76.44 & 75.12 & 72.13 & 70.18 & 76.56 & 73.20 & 73.02 \\
CAF-Mamba~\cite{11465117}     & \textbf{72.17} & \textbf{73.88} & 80.49 & \underline{77.04} & \underline{75.90} & \underline{74.32} & \underline{72.92} & \underline{76.92} & \underline{74.87} & \underline{74.76} \\
% STST~\cite{TAO2024577}           & 70.70 & \underline{72.50} & 77.67 & 75.00 & 73.97 & 67.76     & 69.20     & 64.01     & 66.23     & 66.80     \\
% STCM-Mamba~\cite{11223210}     & \textbf{72.64} & \textbf{73.72} & 82.11 & \textbf{77.69} & \textbf{76.54} & \underline{75.41} & \underline{72.55} & \textbf{81.32} & \underline{76.68} & \underline{76.49} \\
\midrule
%\textbf{Ours}  & \textbf{71.23} & \textbf{70.67} & \underline{86.18} & \textbf{77.66} & \textbf{76.44} & \textbf{76.50} & \textbf{75.00} & \textbf{79.12} & \textbf{77.01} & \textbf{76.91} \\
\textbf{Ours}  & \underline{71.23} & \underline{70.67} & 86.18 & \textbf{77.66} & \textbf{76.44} & \textbf{76.50} & \textbf{75.00} & \textbf{79.12} & \textbf{77.01} & \textbf{76.91} \\
\noalign{\hrule height 1pt}
\end{tabular}
\vspace{-0.5em}
\end{table*}

\subsection{Ablation Study}
\begin{figure}[t]
    \centering
    \begin{subfigure}[b]{0.5\linewidth}
        \includegraphics[width=\textwidth]{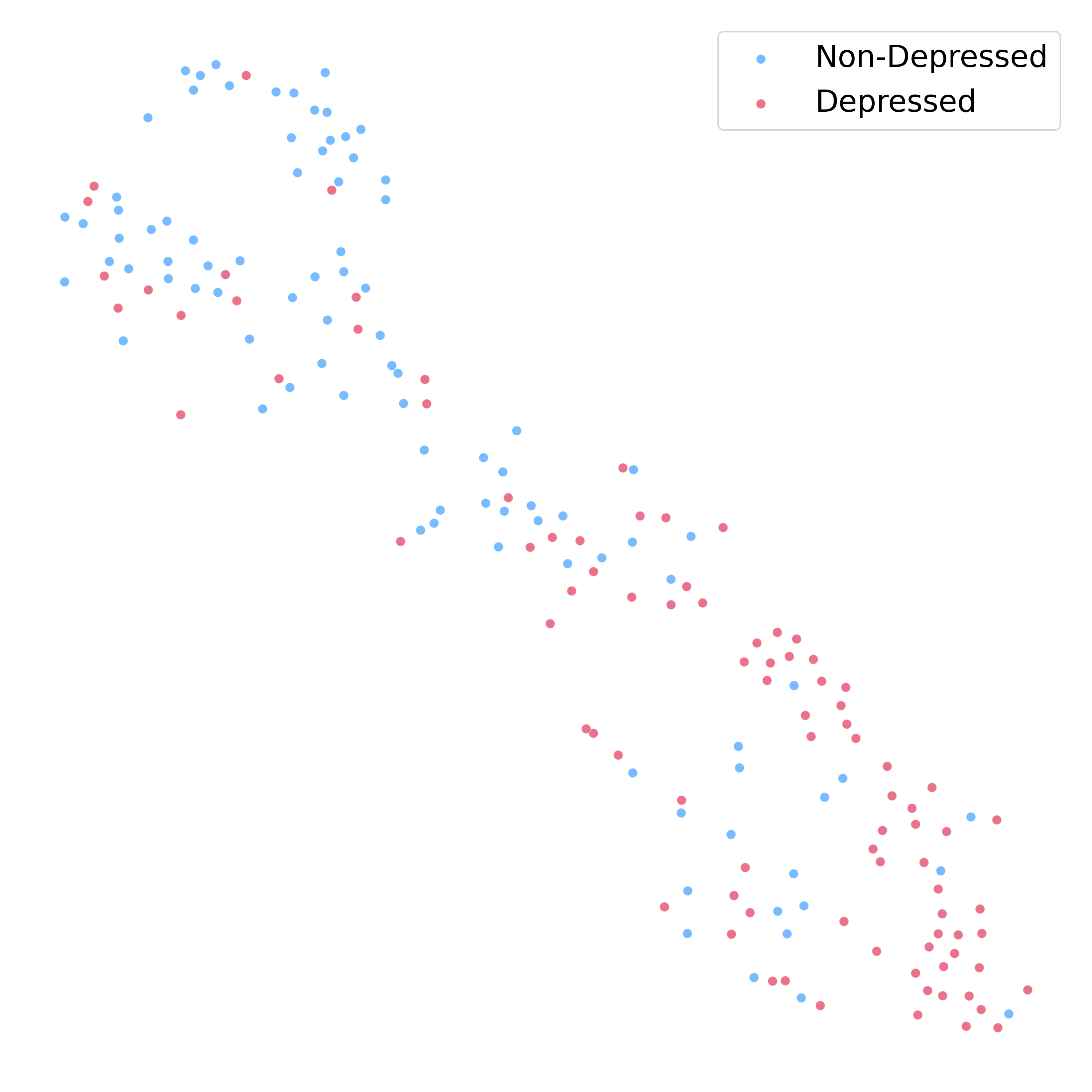}
        \caption{Full model}
        \label{fig:full}
    \end{subfigure}%
    \hfill%
    \begin{subfigure}[b]{0.5\linewidth}
        \includegraphics[width=\textwidth]{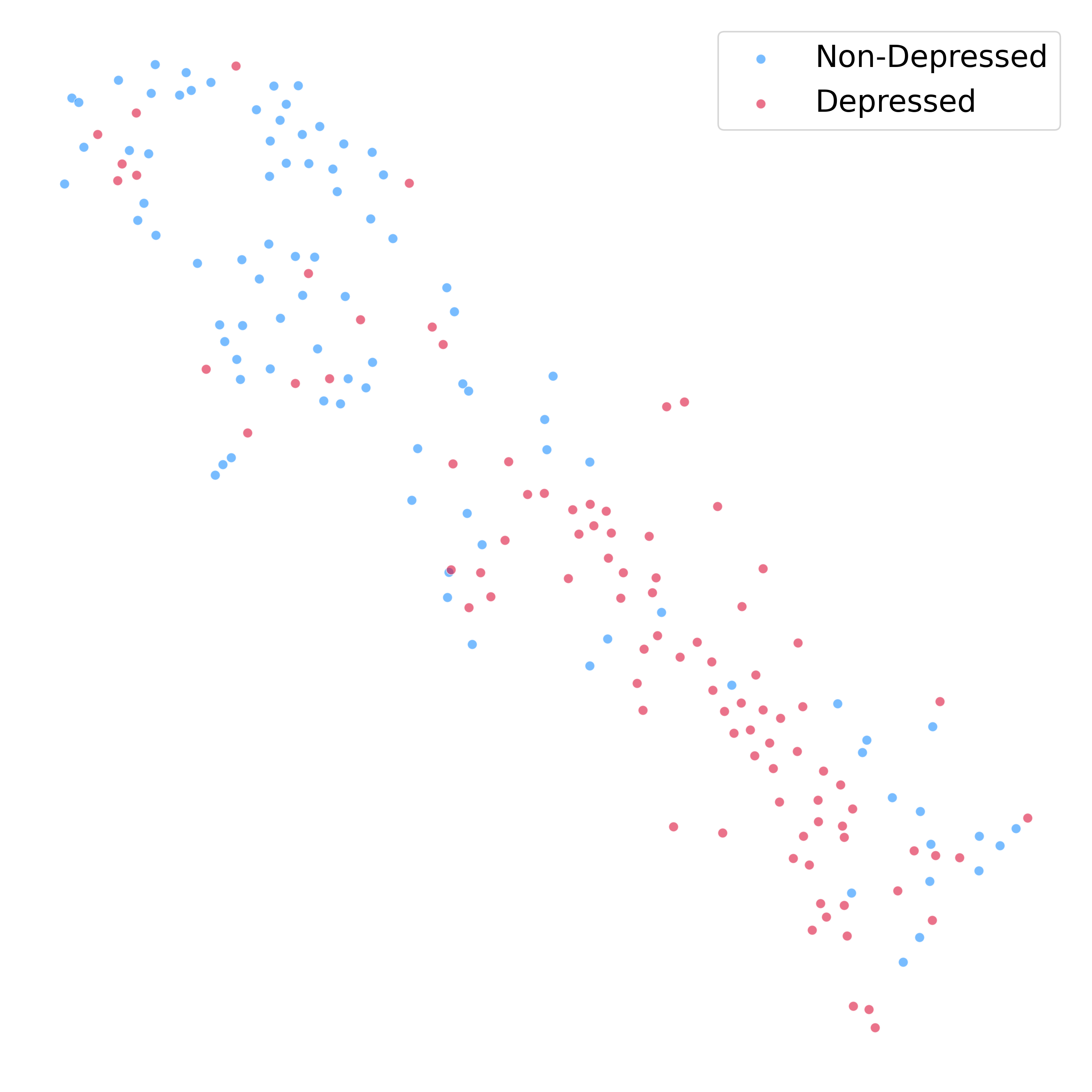}
        \caption{w/o advantage-weighting}
        \label{fig:woadv}
    \end{subfigure}
    \caption{T-SNE visualization of ablation study on LMVD. W/o denotes removing the component.}
    \label{fig:total}
    \vspace{-0.5em}
\end{figure}

\begin{table}[t]
\centering
\caption{Ablation study. W/o denotes removing the component.}
\label{tab:abl}
\begin{tabular}{lcc}
\noalign{\hrule height 1pt}
\textbf{Methods} & \textbf{D-vlog} & \textbf{LMVD} \\
\midrule
w/o mutual transformer (simple concat) & 73.38 & 71.00 \\
w/o advantage-weighting & 70.23 & 73.98 \\
\midrule
\textbf{Full model} & 76.44 & 76.91 \\
\noalign{\hrule height 1pt}
\end{tabular}
\vspace{-0.5em}
\end{table}
% In Figure~\ref{fig:total}, we evaluate decision boundary robustness via the Fisher-like Separation Ratio, defined as $\frac{\text{Inter-class Distance}}{\text{Intra-class Variance}}$, in the 384-D latent space. Without advantage-weighting (Fig. \ref{fig:total}b), severe feature dispersion yields a low ratio of $0.9119$ and a highly overlapped boundary. Conversely, our full model (Fig. \ref{fig:total}a) restrains this dispersion via $\mathcal{L}_{com}$, improving the ratio to $1.0081$. This enhanced signal-to-noise ratio causes t-SNE to project the true class centroids (stars $\star$) further apart, confirming that our framework actively clears the ambiguous boundary zone.

% As shown in Table~\ref{tab:abl}, replacing the mutual transformer with concat or removing the advantage-weighting causes substantial performance drops. The superiority of the full model confirms that fine-grained cross-modal fusion and group-aware ranking are mutually reinforcing.
In Figure~\ref{fig:total}, removing advantage-weighting places more samples near the decision boundary, making the model more susceptible to hard pair misclassification. This vulnerability arises because hard pairs typically occupy regions of high feature overlap where class boundaries are ill-defined, and their proximity to decision surfaces amplifies the impact of minor variations that can flip classification outcomes. Table~\ref{tab:abl} further quantifies these contributions. Removing the mutual transformer leads to a drop in average performance (e.g., -5.91 on LMVD), underscoring the importance of deep cross-modal interaction. Similarly, omitting advantage-weighting results in a marked decline (e.g., -6.21 on D-vlog). These results confirm that fine-grained feature fusion and geometry-aware ranking are synergistic and mutually reinforcing.
\subsection{Analysis of Advantage-weighting Mechanism}
\begin{figure}[t]
    \centering
    \includegraphics[width=1.0\linewidth]{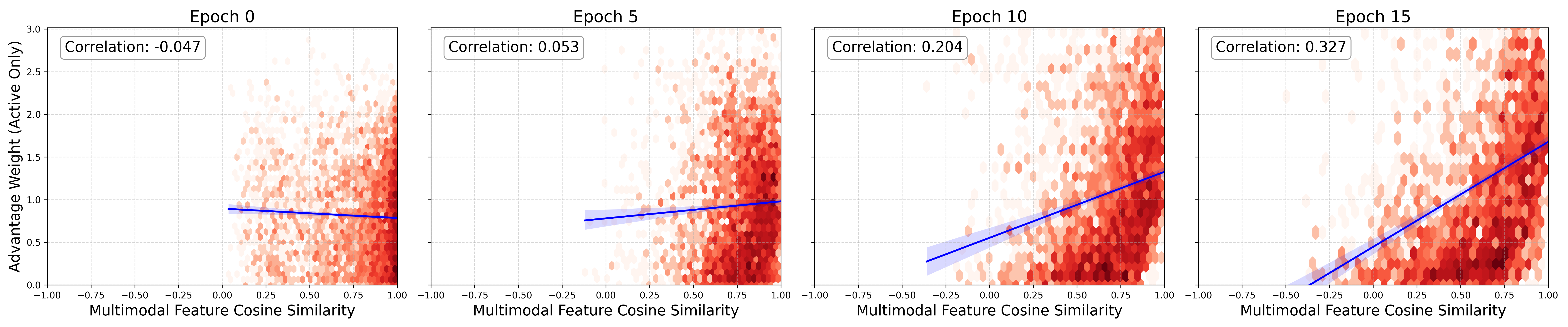}
    \caption{Evolution of advantage-weighting during training. The blue solid line represents the linear regression trend line.}
    \vspace{-0.6em}
    \label{fig:evolution}
\end{figure}

\begin{figure}[t]
    \centering
    \includegraphics[width=0.65\linewidth]{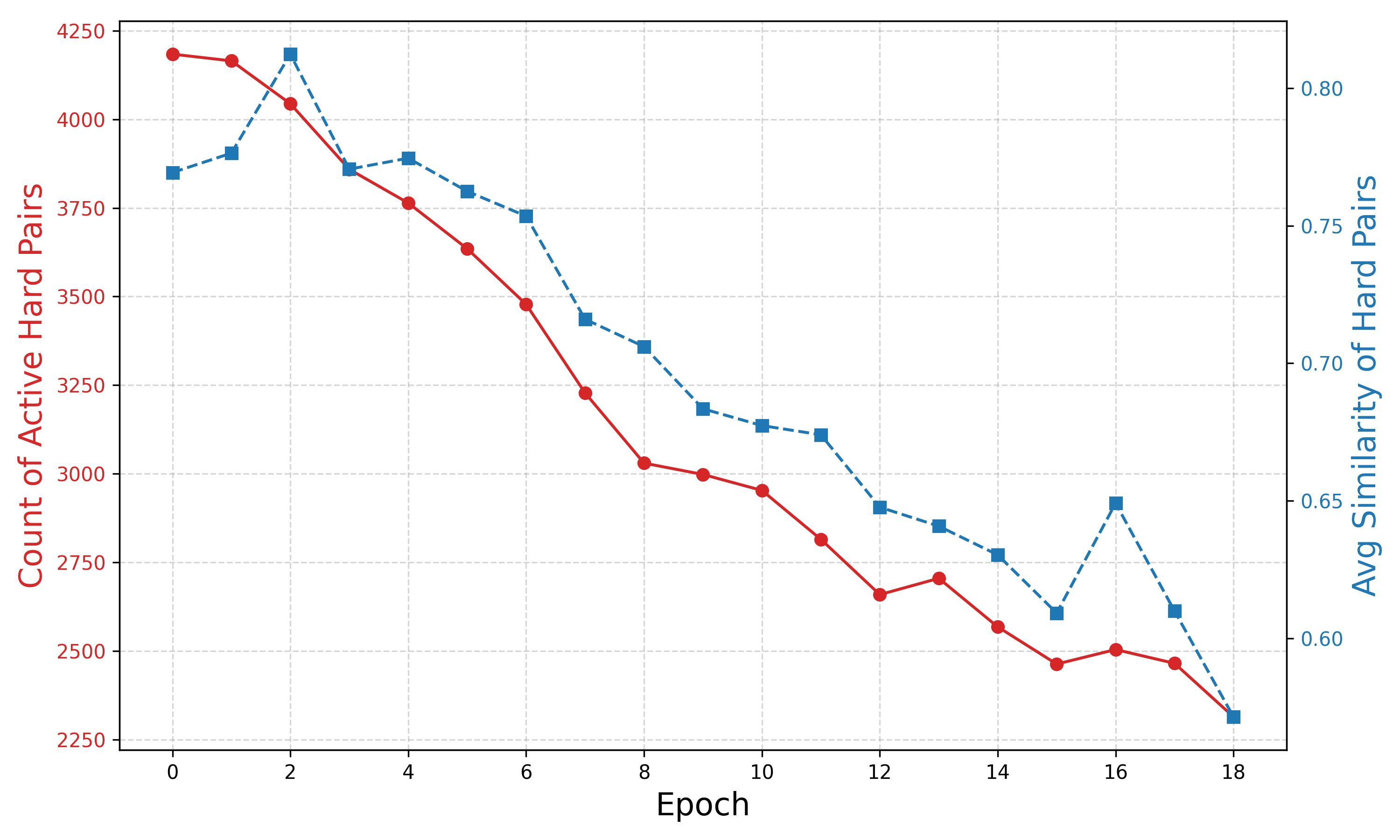}
    \caption{Training dynamics of hard pairs.}
    \vspace{-0.75em}
    \label{fig:training_stats}
\end{figure}
To validate our proposed advantage-weighting mechanism, which uses multimodal feature similarity to identify hard pairs, we visualize how training dynamics evolve. We focus on cross-class pairs with highly similar multimodal features, as these are inherently difficult for the model to distinguish. Throughout training, we track the relationship between the cosine similarity of such pairs and the advantage weights assigned by $\mathcal{L}_{BAR}$.

Figure~\ref{fig:evolution} shows how the advantage weight distribution evolves. Initially, the correlation with similarity of features is near zero ($r=-0.047$), since the randomly initialized model cannot effectively distinguish dissimilar pairs, leading to random weight assignments across similarity values. As training progresses, a clear positive correlation emerges. By epoch 15, the Pearson correlation coefficient increases significantly to $0.327$. The linear regression trend line demonstrates a steep positive slope, indicating that $\mathcal{L}_{BAR}$ dynamically focuses its intensity on pairs with high multimodal similarity. This empirically confirms our hypothesis: as the model resolves easy patterns, the remaining hard pairs are predominantly those with entangled multimodal features, which our mechanism correctly identifies and penalizes.

Figure~\ref{fig:training_stats} provides a quantitative view of this learning process. We monitored the count of active hard pairs (with non-zero advantage weights) and their average cosine similarity against the stable baseline of total pairs ($\approx$ 8,300). Two key trends illustrate the effectiveness. First, the number of active hard pairs drops significantly from 4,184 to 2,463 during training. This indicates that the mechanism successfully resolves ambiguous boundaries for a majority of samples. Second, this reduction in quantity is accompanied by a qualitative shift in the feature space. The average cosine similarity of the remaining hard pairs decreases steadily from 0.77 to 0.57. The joint decline serves as strong evidence that the advantage-weighting mechanism effectively pushes the latent representations of conflicting classes apart, rather than merely filtering out easy ones.

Overall, the advantage-weighting mechanism identifies high-similarity pairs as hard ones, actively optimizes the feature manifold to reduce this similarity, and thus enhances the discriminative power of the multimodal representations.
\subsection{Hyperparameter Stability Experiments}
We investigate the impact of the separation margin $m$ in Table~\ref{tab:hyper_sta}, as Optuna's~\cite{DBLP:journals/corr/abs-1907-10902} parameter importance analysis identified it as the most influential hyperparameter. The model achieves optimal performance at $m=1.15$, forming an inverted U-shaped curve. This value aligns with the optimal trial found by our automatic hyperparameter search. Lower margins ($m=0.75$) provide insufficient discriminative power, while higher margins ($m=1.5$) enforce overly strict constraints that degrade feature learning. The optimal value of $1.15$, identified via Optuna~\cite{DBLP:journals/corr/abs-1907-10902}, balances these factors and is used for the LMVD dataset.

\begin{table}[t]
\centering
\caption{Hyperparameter stability experiments on LMVD.}
\label{tab:hyper_sta}
\begin{tabular}{lccccc}
\noalign{\hrule height 1pt}
\textbf{$m$} & \textbf{Accuracy} & \textbf{Precision} & \textbf{Recall} & \textbf{F1} & \textbf{Average} \\
\midrule
% 0.0 & 0 & 0 & 0 & 0 & 0 \\
% 0.5 & 75.41 & 76.74 & 72.53 & 74.58 & 74.82 \\
0.75 & 72.13 & 71.74 & 72.53 & 72.13 & 72.13 \\
1.0 & 71.04 & 70.21 & 72.53 & 71.35 & 71.28 \\
\textbf{1.15} & 76.50 & 75.00 & 79.12 & 77.01 & 76.91 \\
1.25 & 75.41 & 73.47 & 79.12 & 76.19 & 76.05 \\
1.5 & 73.22 & 73.33 & 72.53 & 72.93 & 73.00 \\
% 2.0 & 77.05 & 78.16 & 74.73 & 76.40 & 76.59 \\
% 2.5 & 70.49 & 70.33 & 70.33 & 70.33 & 70.37 \\
% 3.0 & 72.68 & 74.12 & 69.23 & 71.59 & 71.91 \\
\noalign{\hrule height 1pt}
\end{tabular}
\vspace{-0.5em}
\end{table}

\section{Conclusion and Future Work}
We presented a fine-grained multimodal framework for ADD that bridges the gap between discrete binary labels and the continuous spectrum of depression severity. Our key contribution, the BAR Loss, effectively mitigates the issues of reward sparsity and feature overlap by explicitly modeling ordinal rankings and focusing on hard examples. Experimental results show that our method outperforms baselines on the LMVD dataset and achieves comparable performance on D-vlog. This study highlights the potential of pairwise learning and advantage-weighting in recovering latent ordinal information, offering a robust solution for non-invasive mental health screening.

Our evaluation currently focuses on in-the-wild datasets. Future work will test BAR Loss on clinical datasets (e.g., DAIC-WoZ~\cite{gratch2014distress}) to verify its generalizability and robustness against domain shifts from social media to clinical settings.% Moreover, we plan to validate the generalizability of the advantage-weighting mechanism on Mamba-based multimodal model architecture for fair comparison.% access its robustness against significant domain shifts between social media vlogs and clinical interactions.

\section{Acknowledgment}
This work is supported by the Guangdong Philosophy and Social Sciences Planning Project (No. GD26YJY34).

\section{Generative AI Use Disclosure}
In the preparation of this manuscript, the authors used generative AI tools (e.g., Gemini) to polish the text, instead of producing a significant part of the manuscript.
% \section{Acknowledgments}
% This work was supported by 

% \ifcameraready
%      The Interspeech 2026 organizers
% \else
%      The authors
% \fi
% are responsible for the content and conclusions.

\bibliographystyle{IEEEtran}
\bibliography{mybib}

\end{document}